\documentclass[11pt,a4paper]{article}
\usepackage[hyperref]{naaclhlt2019}
\usepackage{times}
\usepackage{latexsym}
\usepackage{bbm}
\usepackage{graphicx}
\usepackage{mathrsfs}
\usepackage{amsfonts}
\usepackage{amsmath}
\usepackage{url}
\usepackage{CJKutf8}
\aclfinalcopy 

\title{Accelerating Transformer Decoding via a Hybrid of Self-attention and Recurrent Neural Network}

\author{Chengyi Wang$^1$ 
  Shuangzhi Wu$^2$ 
  Shujie Liu$^3$ \\
  $^1$Nankai University, Tianjin, China \\
  $^2$Harbin Institute of Technology, Harbin, China \\
  $^3$Microsoft Research Asia, Beijing, China \\
  {\tt cywang@mail.nankai.edu.cn}
  {\tt \{v-shuawu, shujliu\}@microsoft.com}\\}

\date{}

\begin{document}
\begin{CJK*}{UTF8}{gbsn}
\maketitle
\begin{abstract}
  Due to the highly parallelizable architecture, Transformer is faster to train than RNN-based models and popularly used in machine translation tasks. However, at inference time, each output word requires all the hidden states of the previously generated words, which limits the parallelization capability, and makes it much slower than RNN-based ones. In this paper, we systematically analyze the time cost of different components of both the Transformer and RNN-based model. Based on it, we propose a hybrid network of self-attention and RNN structures, in which, the highly parallelizable self-attention is utilized as the encoder, and the simpler RNN structure is used as the decoder. Our hybrid network can decode 4-times faster than the Transformer. In addition, with the help of knowledge distillation, our hybrid network achieves comparable translation quality to the original Transformer.
\end{abstract}

\section{Introduction}
Recently, the Transformer \cite{DBLP:conf/nips/VaswaniSPUJGKP17} has achieved new state-of-the-art performance for several language pairs.
Instead of using recurrent architectures to extract and memorize the history information in conventional RNN-based NMT model \citep{DBLP:journals/corr/BahdanauCB14}, the Transformer depends solely on attention mechanisms and feed-forward layers, in which, the context information in the sentence is modeled with a self-attention method, and the generation of next hidden state no longer sequentially depends on the previous one. The decoupling of the sequential hidden states brings in huge parallelization advantages in model training.  

However, at inference time, due to its dependence on the whole history of previously generated words at each predicting step and a large amount calculation of multi-head attentions, the Transformer is much slower than RNN-based models. This restricts its application in on-line service, where decoding speed is a crucial factor. We conduct analysis to show time cost of different components of both the Transformer and the RNN-based NMT  model (RNMT) in Table \ref{Cost_Analysis} on 1000 pseudo sentences. All models decode the same length target sentences for a fair comparison.  The RNMT is a standard single layer GRU model with 512 embedding size, 1024-hidden dimension \cite{DBLP:journals/corr/BahdanauCB14}. The Transformer(basic) model follows the basic setting in \cite{DBLP:conf/nips/VaswaniSPUJGKP17}. 30K source- and target- vocabularies are used for all models. 
\begin{table}[] \small
\centering
\begin{tabular}{l|c|c|c}
\hline
Sub-module        & RNMT & Trans.(base) & Trans.(1layer) \\ \hline
-Encoding         &   72.3  &    63.2  & 10.6\\ \hline
-Decoding         &     138.0 &    434.1 & 170.5\\ 
$~~$-Attention       &    43.3&    99.3  &  24.9  \\ 
$~~$-SelfAtt/GRU     &    42.9 &     152.0  &  41.5 \\ 
$~~$-FFN                 &   -    &   86.1  & 19.9\\
$~~$-Softmax         &      40.1 &    46.7  & 45.6\\
$~~$-Others           &     11.7&     50.0  & 38.6 \\ \hline
Total             &     210.3 &    497.3  & 181.1\\ \hline
\end{tabular}

\caption{\label{submodule_comp}Time breakdown of RNMT decoding and Transformer decoding, both with beam size 8. Trans.(base) refers to the base Transformer with 6-layer encoder and decoder and Trans.(1layer) refers to the Transformer with 1-layer encoder and decoder. The time is measured on a single Tesla K40 GPU. The ``Others'' item in decoding contains beam expansion, data transmission etc.. }\label{Cost_Analysis}
\end{table}
As shown in Table \ref{Cost_Analysis}, though the Trans.(base) has much deeper encoder, it is still faster than a single layer RNN due to the high parallelizable multi-head attention. However, the decoding cost of the Transformer decoder is a significant issue which is over 3 times of that of RNN decoder and occupies 88\% of the total decoding time. This is dominated by the high frequency of computing target-to-source attention, target self-attention and feed-forward network. We also analyze a single layer self-attention decoder to compare with the single layer RNN and find that even with a big sacrifice of translation quality, self-attention still slower than RNN in decoding.

\begin{figure}
\centering
  \includegraphics[height=0.4\textwidth]{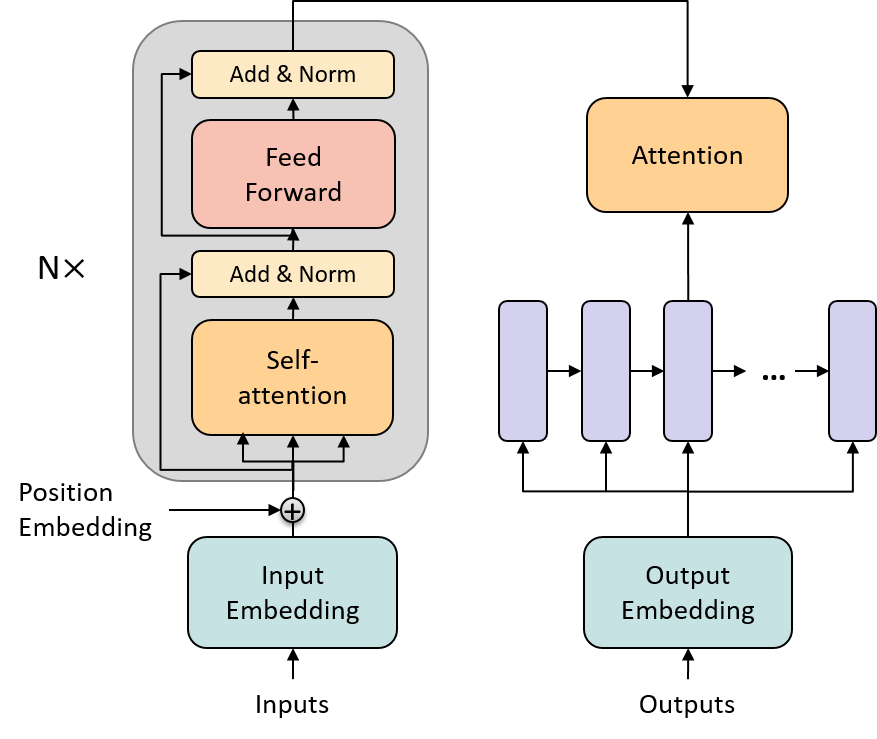}
  \caption{The hybrid architecture.}
  \label{hybrid model}
\end{figure}

As for NMT inference speedup, numerous approaches have been proposed for RNN-based NMT models \cite{devlin:2017:EMNLP2017,zhang2017towards,DBLP:conf/emnlp/KimR16,DBLP:conf/acl/ShiK17}. For Transformer, \citet{gu2017non} proposed a non-autoregressive Transformer where output can be simultaneously computed. They achieved a big improvement on decoding speed at the cost of the drop in translation quality. Recently, \citet{zhang2018accelerating} proposed an efficient average attention to replace the target self-attention of decoder. However, this mechanism cannot be applied to the target-to-source multi-head attention and cannot reduce the calculation of FFN which are the bottleneck for further speedup as shown in Table \ref{Cost_Analysis}.

In this paper, we propose a hybrid architecture where the self-attention encoder and RNN decoder are integrated. Both the two modules are fast enough as shown in Table \ref{Cost_Analysis}. By replacing the original Transformer decoder with an RNN-based module, we speed up the decoding process by a factor of four. Furthermore, by leveraging the knowledge distillation\cite{DBLP:journals/corr/HintonVD15,DBLP:conf/emnlp/KimR16}, where the original Transformer is regarded as the teacher and our hybrid model as the student, our hybrid model can improve the translation performance significantly, and get comparable results with the Transformer. 

\section{Our Hybrid Model}
\subsection{Model Architecture}
As shown in Figure \ref{hybrid model}, our hybrid model contains a self-attention based encoder and an RNN based decoder. In this section, we describe the details of our hybrid model.

\paragraph{Encoder} For our hybrid model, the encoder stays unchanged from the original Transformer network, which is entirely composed of two sub-layers: self-attention modules and feed-forward networks. The self-attention is a multi-head attention network which generates the current state by integrating the whole source context. The following feed-forward layer is composed of two linear transformations with a ReLU activation in between. The layer normalization and residual connection are used after each sub-layer. To model position information,  additive position embeddings are used. 

This kind of encoder avoids the recurrence in RNNs. The self-attention connects all positions with a constant number of operations and each position has direct access to all positions, which ensures the model to learn more distant temporal dependencies. 
Unlike the RNN encoder processing the sentence word by word using sequential operations as long as the sentence, self-attention network only depends on the output of the previous layer, no need to wait for hidden states to propagate across time, which improves model parallelization and speeds up both the training and inference process. Based on the above analysis, self-attention network is leveraged as the encoder of our hybrid mode.

\paragraph{Decoder} 

The original Transformer decoder contains three sub-modules in each layer: an inner self-attention between the current state and target history states, an inter multi-head attention between target state and source context, and a feed forward network. 
This structure is highly parallelizable in training, however, during inference, it is impossible to take full advantage of parallelism because target words are unknown. It has to generate target words one after another as RNN does. In addition, the inner self-attention must access to all history states, which increase the  complexity, and the inter multi-head attention is computed in each layer with the same computational complexity with the inner self-attention as shown in Table  \ref{Cost_Analysis}. As for the RNN decoder, it predicts each target word just depending on the previously hidden state, the previous word and the source context computed by the attention mechanism. From Table \ref{Cost_Analysis}, we can see that the Transformer decoder is the most computationally expensive part which is almost 3 times slower than RNMT, so we leverage a single layer RNN decoder with GRU \cite{DBLP:conf/emnlp/ChoMGBBSB14} as recurrent unit for our hybrid model. It notes that the network structure of \citet{chen2018best} is a little similar to ours. But they focus on finding an optimal structure to improve the translation quality and combine the encoder and decoder from different model families. Their way of combination keeps a larger amount of model parameters, while we aim at accelerating the decoding speed with light network.

\paragraph{Attention mechanism}
To find the most suitable attention functions, we use three different attention functions: additive attention \cite{DBLP:journals/corr/BahdanauCB14}, dot-product attention and multi-head attention. The multi-head attention is as same as the one in our encoder without FFN layer. 

\subsection{Model Training}

We use a two-stage training for our hybrid model: the pre-training and the knowledge distillation fine-tuning. In the first stage, our model is generally trained to maximize the likelihood estimation using the bilingual training data. In the second stage, we apply sequence-level knowledge distillation \textbf{KD} \cite{DBLP:journals/corr/HintonVD15,DBLP:conf/emnlp/KimR16} where the Transformer is regarded as the teacher model and our hybrid model is the student model. Formally, given the bilingual corpus \( D = \{ (x^{(n)}, y^{(n)}) \}_{n=1}^{N} \) where $x$ is the source sentences and  $y$ is the corresponding target ones, the training objective of the second stage is:
\begin{equation}
\begin{aligned}
L({\theta}_s) & = \sum_{n=1}^{N} \log{ P(y^{(n)}| x^{(n)};{\theta}_s)} \\
& - \lambda \sum_{n=1}^{N} \text{KL}( P(y|x^{(n)}; {\theta}_t) || P(y|x^{(n)}; {\theta}_s)) 
\end{aligned}
\label{regularization}
\end{equation}where \( \lambda \) is a hyper-parameter for regularization terms which is set to 1 in all experiments, ${\theta}_s$ and ${\theta}_t$ are model parameters of the student and teacher models, and KL is the Kullback-Leibler divergence terms. ${\theta}_t$ is pre-trained and fixed. In Equation \ref{regularization}, the first term guides the model to learn from the training data and the second term guides it to learn from the teacher network.

\begin{table*}[] \small
\centering
\begin{tabular}{l|c|c|c|c|c|c|c}\hline
  System                        & N & Time & Speed (w/s) & NIST2005 & NIST2008 & NIST2012 & Avg.\\ \hline
  \hline
  RNMT                          & 1-1 & 279s   & 441.2& 39.64 & 30.93 & 29.20 & 33.26   \\ 
  +KD                           & 1-1  & 278s    & 441.8 & 41.25 & 32.90 & 31.24 & 35.13   \\ \hline\hline
  Trans.(teacher)               & 4-4 & 652s & 187.9 & 44.13 & 35.55 & 33.19 & 37.62   \\ \hline
  Hybrid+additive\_attn         & 4-1 &\bf{233s} & \bf{524.2} & 42.10 & 33.87 & 31.80 & 35.92     \\ 
  +KD                           & 4-1  &\bf{232s}  & \bf{526.3}& 43.75 & 35.24 & 33.30 & 37.43        \\\hline
  Hybrid+dot\_attn              & 4-1  &  234s & 522.8 & 42.25 & 32.81 & 30.66 & 35.24 \\ 
  +KD                           & 4-1   &234s  & 523.1 & 44.02 & 34.91 & 33.12 & 37.35 \\\hline
  Hybrid+multi\_attn            & 4-1  & 235s & 523  & 42.82 & 33.73 & 31.95 & 36.17  \\ 
  +KD                           & 4-1  & 236s & 521.1 & 44.05 & 35.46 & 33.29 & 37.60  \\\hline\hline
  Trans.(teacher)               & 6-6 & 1020s & 120.9& 44.64 & 36.01 & 34.19 & 38.28   \\ \hline
  Hybrid+additive\_attn         & 6-1  &250s & 495.8 & 42.65 & 33.93 & 31.75 & 36.11   \\ 
  +KD                           & 6-1  &252s  & 492.3& 44.35 & 36.10 & 33.91 & 38.12   \\ \hline
  Hybrid+dot\_attn              & 6-1  &\bf{249s}  & \bf{491.8}& 43.08 & 32.83 & 31.23 & 35.71 \\ 
  +KD                           & 6-1  &\bf{250s}  & \bf{492}& 44.71 & 35.35 & 33.34 & 37.80    \\ \hline
  Hybrid+multi\_attn            & 6-1 &253s &  486.4& 44.08 & 34.31 & 31.98 & 36.79  \\
  +KD                           & 6-1  &251s & 490.4& 44.88 & 36.04 & 33.80 & 38.24 \\ \hline

\end{tabular}

\caption{Decoding time and case-insensitive BLEU scores (\%) for Chinese-English  NIST datasets. ``Trans.'' is short for Transformer model. ``N'' refers to the depth of encoder and decoder. ``Time" refers to the total decoding time on all the testsets and ``Speed(w/s)" denotes the decoding speed measured by word per second. The
``Avg." is short for the average BLEU score. 
  }\label{Nist}
\end{table*}

\section{Experiment}
\subsection{Setup}
Our proposed model is evaluated on NIST OpenMT Chinese-English and WMT 2017 Chinese-English tasks. All experimental results are reported with IBM case-insensitive BLEU-4 \cite{DBLP:conf/acl/PapineniRWZ02} metric.

\textbf{Dataset:}  
For NIST OpenMT's Chinese-English task, we use a subset of LDC corpora, which contains 2.6M sentence pairs. NIST 2006 is used as development set and NIST 2005, 2008, 2012 as test sets. We keep the top 30K most frequent words for both sides, and others are replaced with \verb|<unk>| and post-processed following \citet{DBLP:conf/acl/LuongSLVZ15}.
For WMT 2017 Chinese-English task, we use all available parallel data, which consists of 24M sentence pairs\footnote{http://www.statmt.org/wmt17/translation-task.html}. 
The newsdev2017 is used as development set and the newstest2017 as the test set.  All sentences are segmented using byte-pair encoding(BPE) \cite{DBLP:conf/acl/SennrichHB16a}. 46K and 33K tokens are adopted as source and target vocabularies respectively.  

\textbf{Baselines and Implement details:}
We compare the decoding speed of our model with a standard 1-layer RNN-based NMT model (RNMT) \cite{DBLP:journals/corr/BahdanauCB14} and a base Transformer model \cite{DBLP:conf/nips/VaswaniSPUJGKP17}. Both baselines are implemented with pre-computing and weight combination techniques.
Specifically, for RNMT, we use precomputed attention \cite{devlin:2017:EMNLP2017} and for Transformer, we pre-compute $(K, V)$ of inter multi-head attention and cache all previous computed $(K, V)$ in inner self-attention for each layer. Linear operations in RNN or multi-head attention are combined into one. For Transformer, the model size is 512 and FFN filter size is 2048. Two different Transformer systems are used: one is 8 heads + 4 encoder/decoder layers, the other is 8 heads + 6 encoder/decoder layers.
RNMT is a single layer GRU model with 512 embedding size, 1024-hidden dimension and additive attention. Our hybrid model uses the same encoder parameters with Transformer and same decoder parameters with RNMT. All the experiments are conducted on a single TITAN Xp GPU. 
In inference, beam search is used with a size 12 and a length penalty 1.0. The decoding batch size is set to 1 for all the experiments.

\subsection{Results and Discussions}
Table \ref{Nist} shows the decoding speed and BLEU scores of different models on NIST datasets. For each hybrid model, we use Transformer model with corresponding layers as the teacher. From the Table \ref{Nist}, all our hybrid models are faster than Transformer and the 1-layer RNMT model. Specifically, our 4- and 6-layer hybrid models achieve significant speedup with factors of 2.8x and 4.1x compared with the 4-layer and 6-layer Transformer teachers. We can find that the time cost of the three different attention models is very close. This is mainly due to the pre-computation and weight combination. As for translation performance, Both Transformers and the hybrid models outperform the RNMT and RNMT+KD.  With the help of sequence-level knowledge distillation, all the hybrid models achieve significant improvements and even get comparable results with the Transformer. 

We further verify the effectiveness of our approach on WMT 2017 Chinese-English translation tasks. Results are shown in Table \ref{WMT}. Similar with the above results, our hybrid models can get 2.3x and 3.9x speedup compared with 4-layer and 6-layer Transformer, and with help of the knowledge distillation, our models achieve comparable BLEU scores with the Transformer. 

Our offline experiments show that the hybrid model improves when the RNN decoder becomes deeper (2, 4, 6 layers), but with slower decoding speed. However, after applying KD, they get similar BLEUs as the hybrid model with single layer decoder. 

\begin{table}[] \small
\centering
\begin{tabular}{l|c|c|c|c}\hline
  System            & N & Time & Speed(w/s) & test\\ \hline
  \hline
  RNMT              & 1-1 & 114s& 500.6 & 20.03 \\
  +KD               & 1-1   & 113s & 505 & 21.03 \\ \hline\hline
  Trans.(teacher)  & 4-4 &234s & 238.1 & 22.57\\ \hline
  Hybrid+dot\_attn  & 4-1  &\bf{100s} & \bf{554.8} & 21.43 \\ 
  +KD               & 4-1  &\bf{99s} & \bf{560.5} & 22.96 \\\hline
  Hybrid+multi\_attn   & 4-1  &105s & 540.5 & 22.05 \\ 
  +KD               & 4-1  &104s & 545.5    & 22.35\\\hline \hline
  Trans.(teacher) &6-6  &420s &130.2   & 23.08\\ \hline
  Hybrid+dot\_attn   & 6-1    &\bf{109s} & \bf{510.2} & 21.63 \\ 
  +KD                &  6-1  &\bf{107s} & \bf{519.9} & 22.50\\ \hline
  Hybrid+multi\_attn  & 6-1  &113s &500.0  & 21.97\\ 
  +KD               &   6-1 &114s & 495.7 & 22.93 \\ \hline

\end{tabular}
\caption{Decoding time and case-sensitive BLEU scores (\%) for WMT 2017 Chinese-English task.}\label{WMT}
\end{table} 




\section{Conclusion}
In this work, we propose a hybrid network to accelerate Transformer decoding. Specifically, we use a self-attention network as the encoder and a single layer RNN as the decoder. Our hybrid models fully take advantage of the parallelization capability of self-attention and the fast decoding ability of RNN-based decoder. In addition, to improve the translation quality, we firstly pre-train our model using the MLE-based method, and then the sequence-level knowledge distillation is used to fine tune it. Experiments conducted on Nist and WMT17 Chinese-English tasks show that our hybrid network gains significant decoding speedup, and achieves comparable translation results with the strong Transformer baseline. 
\bibliography{naaclhlt2019}
\bibliographystyle{acl_natbib}

\appendix

\end{CJK*}
\end{document}